\pdfoutput=1

\documentclass[11pt]{article}

\usepackage[final]{acl}

\usepackage{times}
\usepackage{latexsym}
\usepackage{amsfonts}
\usepackage{amsmath}
\usepackage{amssymb}
\usepackage{balance}

\newcommand{\mX}{\mathbf{X}}

\newcommand{\vphi}{\boldsymbol{\phi}}
\newcommand{\vpsi}{\boldsymbol{\psi}}

\newcommand{\stkout}[1]{{\ifmmode\text{\sout{\ensuremath{#1}}}\else\sout{#1}\fi}}

\usepackage[T1]{fontenc}

\usepackage[utf8]{inputenc}

\usepackage{microtype}

\usepackage{inconsolata}
\usepackage{url}
\usepackage{makecell}
\usepackage{multirow}
\usepackage[hang,flushmargin]{footmisc}
\usepackage{graphicx}
\usepackage{arydshln}

\title{ LLMEmbed: Rethinking Lightweight LLM's Genuine Function \\ in Text Classification}

\author{Chun Liu\textsuperscript{1}\thanks{\noindent Equal contribution.}, Hongguang Zhang\textsuperscript{1}\footnotemark[1], Kainan Zhao\textsuperscript{1}, Xinghai Ju\textsuperscript{2} \and Lin Yang\textsuperscript{1}\thanks{\noindent Corresponding author.} \\
\textsuperscript{1}Systems Engineering Institute, AMS, Beijing, China \\
\textsuperscript{2}State Key Laboratory of Mathematical Engineering \\ and Advanced Computing, Zhengzhou, China \\
\texttt{chunliu-cs@outlook.com, zhang.hongguang@outlook.com, yanglin61s@126.com}
}

\begin{document}
\maketitle
\begin{abstract}
With the booming of Large Language Models (LLMs), prompt-learning has become a promising method mainly researched in various research areas. Recently, many attempts based on prompt-learning have been made to improve the performance of text classification. However, most of these methods are based on heuristic Chain-of-Thought (CoT), and tend to be more complex but less efficient. In this paper, we rethink the LLM-based text classification methodology, propose a simple and effective transfer learning strategy, namely LLMEmbed, to address this classical but challenging task. To illustrate, we first study how to properly extract and fuse the text embeddings via various lightweight LLMs at different network depths to improve their robustness and discrimination, then adapt such embeddings to train the classifier. We perform extensive experiments on publicly available datasets, and the results show that LLMEmbed achieves strong performance while enjoys low training overhead using lightweight LLM backbones compared to recent methods based on larger LLMs, \textit{i.e.} GPT-3, and sophisticated prompt-based strategies. {Our LLMEmbed achieves adequate accuracy on publicly available benchmarks without any fine-tuning while merely use 4\% model parameters, 1.8\% electricity consumption and 1.5\% runtime compared to its counterparts. Code is available at: \url{https://github.com/ChunLiu-cs/LLMEmbed-ACL2024}.} 
\end{abstract}

\section{Introduction} \label{sec:intro}
Recently Large Language Models (LLMs) have shown remarkable abilities on various NLP applications, such as GPT \cite{brown2020language, openai2023gpt}, PaLM \cite{chowdhery2023palm} and LLaMA \cite{touvron2023llama_a,touvron2023llama_b}, offering services to users through dialogue. Since LLMs refer to large-scale pre-trained language models (PLMs) that undergo extensive training on massive textual corpora to understand the complexity and relationships within language, LLMs exhibit amazing \textit{emergent abilities} in comprehension and reasoning \cite{wei2022emergent}. This phenomenon further promotes research of prompt learning for LLMs \cite{white2023prompt}, such as Chain-of-Thought (CoT) \cite{wei2022chain} and Tree-of-Thoughts (ToT) \cite{yao2023tree}.

Prompt-learning~\cite{liu2023pre} is closely connected to the training and inference of LLMs. Instead of adapting the pre-trained language models (PLMs) to address downstream tasks, prompt-learning directly adapts LLMs to cloze-style prediction, autoregressive modeling, or sequence to sequence generation, leading to promising performances on various tasks \cite{ding2022openprompt}. Its major advantage of is that, given a suite of appropriate prompts, LLMs can be used to solve a great number of tasks \cite{brown2020language, sun2021ernie} with no necessity of training from the scratch. However, such a paradigm is deeply involved with prompt engineering to find the most appropriate prompts to improve the overall performance.

Based on the sophisticated prompts, LLMs are guided to generate results and have achieved counterpart, or even stronger performance comparable to supervised baselines in various downstream NLP tasks such as natural language inference \cite{sheng2023flexgen}, question answering \cite{trivedi2023interleaving}, information extraction \cite{josifoski2023exploiting}, named entity recognition \cite{wang2023gpt} and relation extraction \cite{wadhwa2023revisiting}. Nevertheless, as a generative model, LLMs still underperform discriminative models in text classification. Recently Sun \textit{et al.} introduced the so-called clue and reasoning prompting (CARP) \cite{sun2023text} to guide GPT-3 through text classification and got the SOTA performances on 5 widely-used text classification benchmarks.

\begin{figure*}[t]
    \centering
    \includegraphics[width=\linewidth]{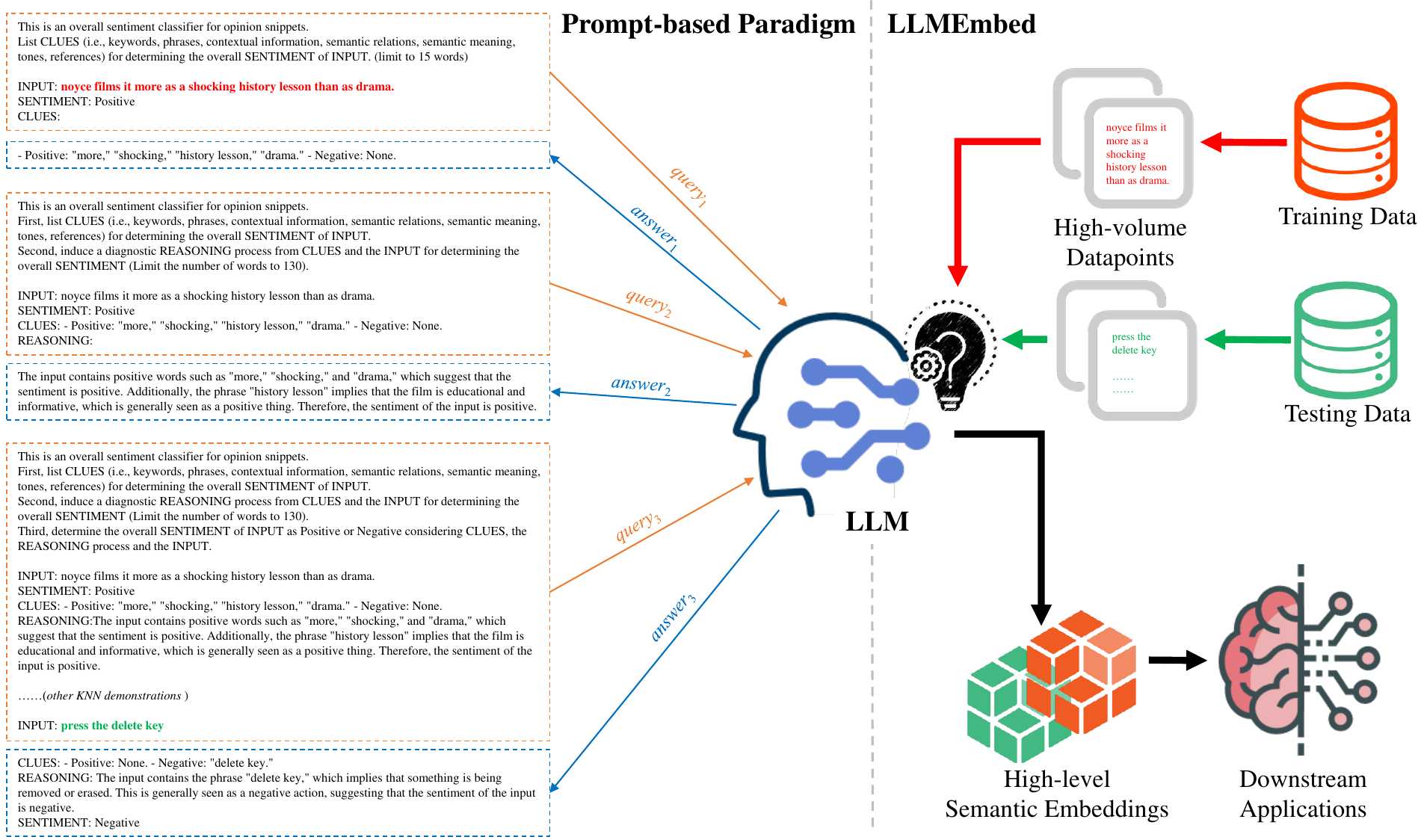}
    \caption{The principle of our proposed LLMEmbed. The left part shows how recent prompt-based methods work to classify texts. It can be seen that such a multi-step reasoning process can merely be performed serially, thus leading to high inference overhead. For comparison, instead of using LLM's content generation ability, we use the latent semantic embeddings extracted by LLMs to realize a much more effective adaptation for downstream classification tasks.}
    \label{fig:enter-label}
\end{figure*}

For specific scenario customization of LLMs \cite{zhang2024tinyllama} or considerations regarding user data privacy security \cite{tan2024beimingwu}, there is still a need for the localized deployment of LLMs. However, deploying LLMs is not always easily available as it requires enormous amounts of computation for training and inference \cite{xia2023sheared}. Moreover, it is worthy noting that only lightweight LLMs (\textit{e.g.} LLaMA2-7B) are open-source (compared to GPT-3 with 175B parameter scale, 7B is lightweight) by far. Subsequently, most users can merely deploy the open-source lightweight LLMs. As the parameter scale of open-source lightweight LLMs is far smaller than online LLMs, the emergent abilities of lightweight LLMs fall behind \cite{wei2022emergent}. 

In this paper, we rethink if the prompt-based paradigm remain effective to lightweight LLMs for text classification, and propose a novel and effective paradigm, namely LLMEmbed to improve the overall training efficiency and generalized performance. Specifically, we fully investigate the effectiveness of prompt-based methodology in text classification, and realize a low-cost and easy-to-use transfer learning framework via the use of LLM embeddings. We perform extensive experiments on publicly available datasets and fairly compare LLMEmbed with recent baselines. It is observed that the the inference of prompt-based paradigms is complex and highly costly. Moreover, their performance is not as promising as CARP \cite{sun2023text} claimed when lightweight LLMs are used, some of the generated results even deviates from the inputs, \textit{i.e.} \textit{hallucination} \cite{zhang2023siren, huang2023survey}. For comparison, our LLMEmbed enjoys a highly robust performance and very low training overhead in all scenarios, which further highlights the usefulness of LLMEmbed. 

In summary, our contributions in this paper are listed as follows:
\begin{itemize}
\item To the best of our knowledge, we are the first to adapt lightweight LLM's semantic embeddings for text classification. We propose a simple but effective paradigm, namely LLMEmbed, based on lightweight LLMs to address the text classification task. Our paradigm achieves SOTA results compared with prompt-based methods with the same lightweight LLM backbone, and comparable performance to methods using large-scale LLMs.
\item Our LLMEmbed paradigm directly constructs the mapping from input texts to output classification results. Therefore, there is no need for users to design sophisticated prompts to align inputs and outputs, \textit{i.e.} there exists no hallucination. Compared to existing works, LLMEmbed is budget-saving since no extra token overhead is required. 
\item Our LLMEmbed is more flexible, scalable and efficient compared to prompt-based methods. LLMEmbed can combine the embeddings of lightweight LLMs with discriminative models (\textit{e.g.} RoBERTa and BERT), or employ other representation learning methods to improve the classification performance. Moreover, LLMEmbed can perform classification in a high-speed parallel manner by feeding a large batch of text datapoints as input, whereas prompt-based paradigms cannot.
\end{itemize}

\section{Related Work}

\subsection{Lightweight Large Language Models}
Scaling up the parameters of a language model has demonstrated the effectiveness on various NLP tasks \cite{brown2020language,chowdhery2023palm}. Especially, the breakthrough achieved by ChatGPT/GPT-4 \cite{openai2023gpt} has made LLMs a promising approach to understanding language and generating contents. However, considering specific scenario customization of LLMs \cite{zhang2024tinyllama}, user data privacy security \cite{tan2024beimingwu}, and limited computational resource \cite{xia2023sheared}, open-source lightweight LLMs also aroused researchers' interests, such as LLaMA2-7B \cite{touvron2023llama_b}, OPT-6.7B \cite{zhang2022opt}, Pythia-6.9B \cite{biderman2023pythia}.

\subsection{Prompt-learning}
Prompt works as the input to guide LLMs to generate contents satisfying users' expectations. With the recent booming of generative LLMs, the prompt engineering \cite{liu2023pre}, which focuses on designing effective prompts, has been a promising research topic. 
\textit{In-Context Learning} (ICL) is a kind of typical prompt-learning which lists examples of the dataset as demonstrations to the LLM without adjusting the LLM’s network architecture \cite{brown2020language,schick2021s}. Besides the demonstrations, \textit{Instruction-Following} introduces task-describing instructions with the desired responses into the prompt, and then fine-tunes LLMs on the instructional data \cite{yi2019towards,wei2021finetuned,mishra2022cross,ouyang2022training,wang2022super}. \textit{Impersonation} is another technique which make LLMs pretend to be a domain expert when answering a domain-specific question \cite{salewski2023context}.

From the viewpoint of human being’s thinking process when solving complicated tasks, Wei et al. proposes \textit{Chain-of-Thought} (CoT) prompting to make LLMs decompose the problem into intermediate steps and solve each before giving the final answer \cite{wei2022chain}. Inspired by CoT, extensions such as \textit{zero-shot variants} \cite{kojima2022large} and \textit{Auto-CoT} \cite{zhang2022automatic} have been introduced. \textit{Self-consistency} samples multiple reasoning paths and selects the most consistent answer via a vote \cite{wang2022self}. \textit{Least-to-Most} decomposes a given complex problem and solves subproblems iteratively to get the final answer \cite{zhou2022least}. \textit{ReAct} allows the language model to interact with external environments such as Wikipedia to incorporate knowledge to inference \cite{yao2022react}. \textit{Self-refine} makes the LLM generate an initial output and then iteratively provide feedback on the previous output, which is used to revise the output \cite{madaan2023selfrefine}. \textit{Tree of Thoughts} (ToT) generalize CoT to maintain a tree of thoughts containing multiple different steps, enabling the LLM to self-evaluate the best way \cite{yao2023tree}.

\begin{figure*}
    \centering
    \includegraphics[width=\linewidth]{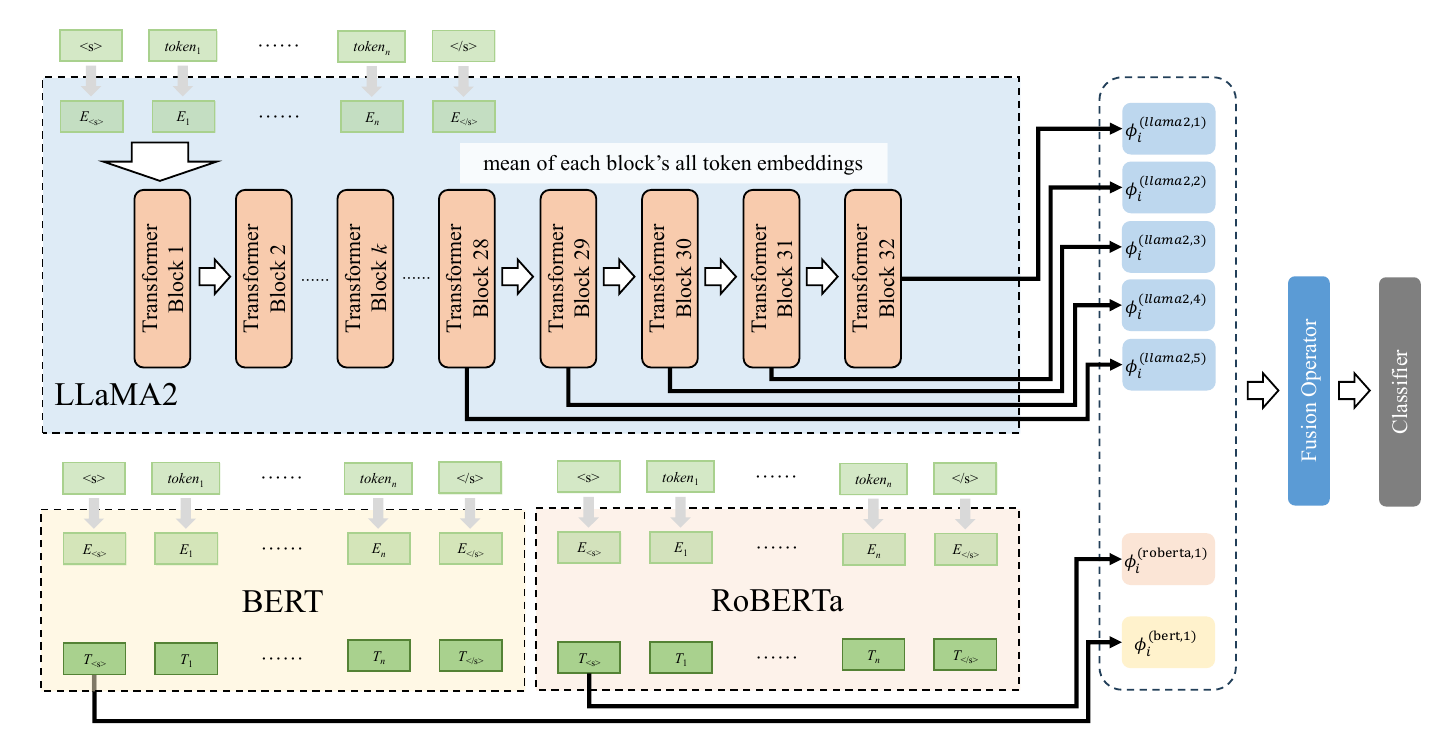}
    \caption{The demonstration of our LLEmbed method. The whole pipeline is a kind of typical transfer learning framework in which the parameters of backbone are pre-trained and frozen, and only the parameters of classifier head is trained during training. We investigate to fuse the semantic embeddings extracted from \textbf{llama2}, \textbf{roberta} and \textbf{bert}. Moreover, for \textbf{llama2}, we extract the embeddings at multiple network depths, and fuse them later via pooling operators to improve the embeddings' generalized ability.}
    \label{fig:pipe}
\end{figure*}

\subsection{Prompt-based Text Classification}
Based on the designed prompt, language models are guided to generate answers as the classification results. Schick et al. reformulated input texts into cloze-style phrases and generated the labels \cite{schick2021exploiting}. Han et al. designed sub-prompts and composed them into final prompts based on logic rules, to guide the language model to generate results \cite{han2022ptr}. Liu et al. designed prompt based on the test sample's semantically similar examples \cite{liu2022makes}, and Shi et al. used k-nearest neighbor (KNN) algorithm to retrieve similar examples as demonstrations of ICL prompts \cite{shi2022nearest}.

Further combining ICL prompts, CoT and KNN algorithm, Sun et al. proposed a method called \textit{clue and reasoning prompting} (CARP) \cite{sun2023text} to guide GPT-3 through text classification and got the SOTA performances on 5 widely-used benchmarks. First, they used texts and labels of training dataset to generate clues and reasoning of text classification by GPT-3. Next, they used PLM such as SimCSE \cite{gao2021simcse} to compute the sentence level representation of training dataset, and then got the most semantically similar demonstrations to the test sample based on KNN algorithm. Finally, based on the reasoning process of these demonstrations, GPT-3 performs reasoning on the test sample and generates the classification result with many rounds of voting.

\begin{table*}[t]
\small
\centering
\caption{Fusion strategies employed in the LLMEmbed. The abbreviations $Avg$, $Max$, $Co$ and $Cat$ refer to average pooling, max pooling, co-occurrence pooling and concatenation, respectively. $v \rightarrow \hat{v}$ represents aligning an embedding vector $v$ to a new embedding vector $\hat{v}$ of new size. Note that the resize operation is executed by a Linear layer, and the sizes of $\vphi_i^{(llama2,d_m)}$, $\vphi_i^{(roberta,1)}$, $\vphi_i^{(bert,1)}$ are 4096-d, 1024-d, 1024-d, respectively.}
\begin{tabular}{c|l|c|l}
    \hline
            \textbf{Index} & \multicolumn{1}{c|}{\textbf{Embeddings}} & \textbf{Operator} $v(\cdot)$ & \multicolumn{1}{c}{\textbf{Description}}  \\
    \hline
        1 & $\vphi_i^{(llama2,1)}$ & / & /   \\
    \hline
        2 & \multirow{3}{*}{$\{\vphi_i^{(llama2, d_m)}\}_{d_m=[1...5]}$} 
        & $Avg$ 
        & $\vpsi_i = Avg(\{\vphi_i^{(llama2, d_m)}\})$ \\ \cline{1-1}\cline{3-4}
        3 &  
        & $Max$ 
        & $\vpsi_i = Max(\{\vphi_i^{(llama2, d_m)}\})$ \\
        \cline{1-1}\cline{3-4}
        4 &  
        & $Cat$ 
        & $\vpsi_i = Cat(\{\vphi_i^{(llama2, d_m)}\})$ \\
    \hline
        5 & \multirow{3}{*}{\makecell[l]{$\{\vphi_i^{(llama2, d_m)}\}_{d_m=[1...5]}$, \\ $\vphi_i^{(bert,1)}$}} 
        & {$Avg$ + $Cat$}
        & $\vpsi_i = Cat(Avg(\{\vphi_i^{(llama2, d_m)}\}), \vphi_i^{(bert,1)})$ \\
        \cline{1-1}\cline{3-4}
        6 &  
        & \makecell[c]{$Max$ + $Cat$} 
        & $\vpsi_i = Cat(Max(\{\vphi_i^{(llama2, d_m)}\}), \vphi_i^{(bert,1)})$ \\
        \cline{1-1}\cline{3-4}
        7 & 
        & \makecell[c]{$Cat$} 
        &  $\vpsi_i = Cat(\{\vphi_i^{(llama2, d_m)}\}, \vphi_i^{(bert,1)})$ \\
    \hline
        8 & \multirow{3}{*}{\makecell[l]{$\{\vphi_i^{(llama2, d_m)}\}_{d_m=[1...5]}$, \\ $\vphi_i^{(roberta,1)}$}} 
        & \makecell[c]{$Avg$ + $Cat$}
        & $\vpsi_i = Cat(Avg(\{\vphi_i^{(llama2, d_m)}\}), \vphi_i^{(roberta,1)})$ \\
        \cline{1-1}\cline{3-4}
        9 &  
        & \makecell[c]{$Max$ + $Cat$} 
        & $\vpsi_i = Cat(Max(\{\vphi_i^{(llama2, d_m)}\}), \vphi_i^{(roberta,1)})$ \\
        \cline{1-1}\cline{3-4}
        10 &  
        & \makecell[c]{$Cat$} 
        &  $\vpsi_i = Cat(\{\vphi_i^{(llama2, d_m)}\}, \vphi_i^{(roberta,1)})$ \\
    \hline
    11 & \multirow{10}{*}{\makecell[l]{$\{\vphi_i^{(llama2, d_m)}\}_{d_m=[1...5]}$, \\ $\vphi_i^{(bert,1)}$, \\ $\vphi_i^{(roberta,1)}$}}
        & {$Avg$ + $Cat$}
        & $\vpsi_i = Cat(Avg(\{\vphi_i^{(llama2, d_m)}\}), \vphi_i^{(bert,1)}, \vphi_i^{(roberta,1)})$ \\
        \cline{1-1}\cline{3-4}
        12 & 
        & \makecell[c]{$Max$ + $Cat$}
        & $\vpsi_i = Cat(Max(\{\vphi_i^{(llama2, d_m)}\}), \vphi_i^{(bert,1)}, \vphi_i^{(roberta,1)})$ \\
        \cline{1-1}\cline{3-4}
        13 & 
        & \makecell[c]{$Cat$}
        & $\vpsi_i = Cat(\{\vphi_i^{(llama2, d_m)}\}, \vphi_i^{(bert,1)}, \vphi_i^{(roberta,1)})$ \\
        \cline{1-1}\cline{3-4}
        14 & 
        & $Cat + Co$
        & \makecell[l]{$\mathbf{1: }\{\vphi_i^{(llama2, d_m)}\}_{5 \times 4096} \rightarrow \{\hat{\vphi}_i^{(llama2, d_m)}\}_{5 \times 1024}$ \\
        $\mathbf{2: }Cat(\{\hat{\vphi}_i^{(llama2,d_m)}\}, \vphi_i^{(bert,1)}, \vphi_i^{(roberta,1)}) \rightarrow {\bf X}_{7 \times 1024}$ \\
        $\mathbf{3: }\vpsi_i = PN(Cat({\bf X}{\bf X}^{\rm T}[1:7]), \sigma)$}  \\
        \cline{1-1}\cline{3-4}
        15 & 
        & \makecell[c]{$Cat + Co$ \\ $+ Avg + Cat$}
        & \makecell[l]{$\mathbf{1: }\{\vphi_i^{(llama2, d_m)}\}_{5 \times 4096} \rightarrow \{\hat{\vphi}_i^{(llama2, d_m)}\}_{5 \times 1024}$ \\
        $\mathbf{2: }Cat(\{\hat{\vphi}_i^{(llama2,d_m)}\}, \vphi_i^{(bert,1)}, \vphi_i^{(roberta,1)}) \rightarrow {\bf X}_{7 \times 1024}$ \\
        $\mathbf{3: }\vpsi_i = Cat(PN(Cat({\bf X}{\bf X}^{\rm T}[1:7]), \sigma), Avg(\{\vphi_i^{(llama2, d_m)}\}))$} \\
    \hline
\end{tabular}
\label{tab:fusion}
\end{table*}

\section{Methodology}
As mentioned in the Section \ref{sec:intro}, improving the performance of locally deployed lightweight LLMs is urgently important, considering specific scenario customization of LLMs \cite{zhang2024tinyllama}, user data privacy security \cite{tan2024beimingwu}, and limited computational resource \cite{xia2023sheared}. However, the emergent abilities of lightweight LLMs fall behind online LLMs due to their far smaller parameter scale \cite{wei2022emergent}, which subsequently limits the effectiveness of lightweight LLMs in prompt-based text classification methods. Therefore, we present the so-called LLMEmbed, which directly employ the lightweight LLMs to extract text embeddings, and improve the overall performance of text classification.

In this paper, we extract and fuse embeddings extracted by multiple backbones at different depths to improve the robustness and generalization. Let us take $f(\cdot|m,d_m)$ to denote the embedding extraction where $m$ is the model and $d_m$ is the network depth.
$g(\cdot|\theta_g)$ refers to the classifier head. $D=\{x_i, y_i\}_{i=1}^N$ is the dataset where $x_i$ is the datapoint, $y_i$ is the label and $N$ is the total number of datapoints. $D$ is divided into $D_{train}$ and $D_{test}$ for training and evaluation. 

During training, {we first randomly sample a batch of datapoints $\{x_i,y_i\}_{i=1}^{N_{b}}\in D_{train}$} where $N_b$ refers to the batchsize, then feed them into $f(\cdot|m,d_m)$ to extract the semantic embeddings $\{\vphi_i^{(m,d_m)}\}_{i=1}^{N_{b}}$. 
\begin{equation}
    \vphi_i^{(m,d_m)} = f(x_i|m, d_m),
\end{equation}
where $m\in \{\textbf{llama2}, \textbf{roberta}, and \; \textbf{bert}\}$ refers to the backbone selected, $d_m$ denotes the embedding is extracted from the last $d_m$-th block of model $m$. {Specifically, $d_m\in\{1...5\}$ for \textbf{llama2} where the embedding of each block is the mean of this block's all token embeddings, $d_m=1$ for \textbf{roberta} and \textbf{bert} where the final sentence semantic is used for embedding fusion.}

Following we fuse all available embeddings via the operator $v(\cdot)$ to get the final semantic embedding $\vpsi_i$ for the $i$-th datapoint.

\begin{equation}
    \vpsi_i = v(\{\vphi_i^{(m,d_m)}\}),
\end{equation}

\begin{table*}
\small
\centering
\caption{The accuracy performance of different settings on 5 publicly available datasets. The last row reports the mean accuracy of each method over benchmarks. The bold results indicate the best performance for each dataset.}
\begin{tabular}{cccccccc}
    \Xhline{1pt}
          & \textbf{Backbone}& \textbf{SST-2} & \textbf{MR}  & \textbf{AGNews} & \textbf{R8}  & \textbf{R52}  & \textbf{Avg. Acc.} \\
    \Xhline{0.75pt}
    \multicolumn{8}{l}{\textbf{PLM Methods}} \\
    \Xhline{0.75pt}
    \cite{kenton2019bert} & BERT-large &  0.8761 &  0.8130 &  0.8216 &  0.8250 &  0.6581  & 0.7988\\
    \cite{liu2019roberta} & RoBERTa-large & 0.9025 &  0.9346  &  0.9441 &  0.9676 &  0.4217 & 0.8341 \\
    \Xhline{0.75pt}
    \multicolumn{8}{l}{\textbf{Prompt-based LLM Methods}} \\
    \Xhline{0.75pt}
    IO \cite{brown2020language}  & LLaMA2 7B &  0.8922 &  0.9065  &  0.7420 &  0.2481  &  0.1869 & 0.5951 \\
    CARP \cite{sun2023text} & LLaMA2 7B & 0.8842 &  0.8394 &  0.8518 &  0.7510 &  0.7305 & 0.8114 \\
    CARP \cite{sun2023text} & GPT-3 175B & 0.9569 & 0.9074 & 0.9525 & 0.9783 & \textbf{0.9627} & 0.9516 \\
    \Xhline{0.75pt}
    \multicolumn{8}{l}{\textbf{LLMEmbed-based Methods}} \\
    \Xhline{0.75pt}
    \multicolumn{8}{c}{\textbf{Embeddings:} $\vphi_i^{(llama2,1)}$ or $\{\vphi_i^{(llama2, d_m)}\}_{d_m=[1...5]}$} \\
    \hdashline
     / & LLaMA2 7B &  0.9518 &  0.9522 &  0.9571 &  0.9794 &  0.9455 & 0.9572 \\
    $Avg$ & LLaMA2 7B & 0.9530 & 0.9545 & 0.9566 & 0.9794 & 0.9533 & 0.9594\\
    $Max$ & LLaMA2 7B & 0.9530   & 0.9517   & 0.9555   & 0.9799   & 0.9486 & 0.9577  \\
    $Cat$ & LLaMA2 7B & 0.9518  & 0.9537 & 0.9359 & 0.9785  & 0.4217 & 0.8483 \\
    \hdashline
    \multicolumn{8}{c}{\textbf{Embeddings:} $\{\vphi_i^{(llama2, d_m)}\}_{d_m=[1...5]}$ and $\vphi_i^{(bert,1)}$} \\
    \hdashline
    $Avg + Cat$ & LLaMA2 7B & 0.9553 & 0.9543 & 0.9554 & 0.9799 & 0.9517 & 0.9593 \\
    $Max + Cat$ & LLaMA2 7B & 0.9530 & 0.9526 & 0.9553 & 0.9794 & 0.9502 & 0.9581\\
    $Cat$ & LLaMA2 7B & 0.9495 & 0.9531 & 0.9364 & 0.9781  & 0.8501 & 0.9334\\
    \hdashline
    \multicolumn{8}{c}{\textbf{Embeddings:} $\{\vphi_i^{(llama2, d_m)}\}_{d_m=[1...5]}$ and $\vphi_i^{(roberta,1)}$} \\
    \hdashline
    $Avg + Cat$ & LLaMA2 7B & 0.9541 & 0.9547 & 0.9562 & 0.9808 & 0.9544 & 0.9600 \\
    $Max + Cat$ & LLaMA2 7B & 0.9537 & 0.9538 & 0.9555 & 0.9794 & 0.9467 & 0.9578\\
    $Cat$ & LLaMA2 7B & 0.9507 & 0.9536 & 0.9491  & 0.9790 & 0.4217 & 0.8508\\
    \hdashline
    \multicolumn{8}{c}{\textbf{Embeddings:} $\{\vphi_i^{(llama2, d_m)}\}_{d_m=[1...5]}$, $\vphi_i^{(bert,1)}$, and $\vphi_i^{(roberta,1)}$} \\
    \hdashline
    $Avg + Cat$ & LLaMA2 7B & 0.9553  & 0.9534  & 0.9574  & 0.9808  & 0.9548 & 0.9603 \\
    $Max + Cat$ & LLaMA2 7B & 0.9541  & 0.9542 & 0.9555 & 0.9799 & 0.9529 & 0.9593\\
    $Cat$ & LLaMA2 7B & 0.9484 & 0.9540 & 0.9505  & 0.9785 & 0.9326 & 0.9528 \\
    $Cat + Co$  & LLaMA2 7B & 0.9553   & 0.9533   & 0.9549   &  0.9794  & 0.8895 &  0.9465 \\
    $Cat + Co + Avg + Cat$  & LLaMA2 7B & \textbf{0.9576}   &  \textbf{0.9549}   & \textbf{0.9583}   &  \textbf{0.9822}  & 0.9568 & \textbf{0.9620} \\
    \Xhline{1pt}
\end{tabular}
\label{tab:acc}
\end{table*}

The fusion operators investigated in this paper are listed in Table~\ref{tab:fusion}. It can be seen that average pooling, max pooling, co-occurrence pooling and concatenation are considered. Due to its complexity, here we take co-occurrence pooling as example to demonstrate the fusion process in detail.

\textbf{Co-occurence pooling.} Given the embedding set $\{\vphi_i^{m,d_m}\}_{i=1}^{N_b}$ for each datapoint $x_i$, we first use linear mappings to project them into the same latent space and align them to the same length $1\times K$, thus making $\vphi_i^{m,d_m}\in \mathbb{R}^{1\times K} \;\; \forall m, d$. Then they are stacked to formulate the $\hat{\vphi}_i\in\mathbb{R}^{H\times K}$ where $H=\sum_{m}d_m$. Following, we calculate their co-occurrence statistics $\psi_i\in\mathbb{R}^{K\times K}$.

\begin{align}
    \vpsi_i &\;= PN(\hat{\vphi}_i \otimes \hat{\vphi}_i^T, \sigma), \\
    \textit{s. t.} \;PN(\mX;&\;\sigma) = \frac{1 - e^{-\sigma \mX}}{1+e^{-\sigma \mX}}=\text{tanh}(2\sigma\mX) \nonumber
\end{align}
where $PN(\cdot)$ is a power normalization function used to balance the power distribution of co-occurrences, $\sigma$ is the hyper-parameter to control the slope of $PN(\cdot)$ function, and the empirical range of value is $[0.1, 0.5]$ in this paper.

Finally, all fused embeddings are fed into the classifier head to get the predictions $\tilde{y}_i=g(\vpsi_i|\theta_g)$, and CrossEntropy loss is performed to update its parameters $\theta_g$.

\begin{equation}
    \arg\min\limits_{\theta_g} \mathcal{L} = - \sum_i y_i\log \tilde{y}_i.
\end{equation}

Overall, our LLMEmbed pipeline can be regarded as a typical transfer learning framework, which focuses on difference fusion strategies on LLM-based embeddings. As the embedding extraction process is fast, our model can be trained and adapted to the target domains with very high efficiency.

\begin{table*}[t]
\footnotesize
\centering
\caption{The runtime of each method. The reported results are formulated by 'hh:mm:ss', which refer to hours, minutes and seconds respectively. (The computational device is Nvidia A100-40G.)
}
\makebox[\linewidth]{
\setlength{\tabcolsep}{0.5em}
\fontsize{9}{9}\selectfont
\begin{tabular}{cccccccc}
    \Xhline{1pt}
    Method & Process & SST-2 & MR & AGNews & R8 & R52 & Avg. Time \\
    \Xhline{0.75pt}
    \multirow{4}{*}{CARP} & clues and reasoning generating & 67:54:13 & 42:52:27 & 133:32:52 & 09:28:47 & 08:33:56 \\
                            & KNN with embed. extracting & 00:24:54 & 00:15:34 & 00:45:11 & 00:02:03 & 00:02:28 \\
                             & inference & 11:12:02 & 138:25:05 & 97:32:00 & 30:37:57 & 36:19:59 \\
        \cline{2-7}
                            & total runtime & 79:31:09 & 181:33:06 & 231:50:03 & 40:08:47 & 44:56:23 & 115:36:01 \\
    \Xhline{0.75pt}
    \multirow{5}{*}{LLMEmbed} & train embed. extraction & 00:13:30 & 05:43:45 & 00:49:04 & 00:10:30 & 00:12:39 \\
                        & test embed. extraction & 00:00:10 & 00:33:21 & 00:03:09 & 00:04:15 & 00:04:57 \\
                        & training ($50$ epochs) & 00:08:30 & 00:05:13 & 00:14:57 & 00:01:08 & 00:01:09 \\
                        & inference & 00:00:01 & 00:00:02 & 00:00:02 & 00:00:01 & 00:00:01 \\
        \cline{2-7}
                        & total runtime & 00:22:11 & 06:22:21 & 01:07:11 & 00:15:54 & 00:18:45 & 01:41:16 \\
    \Xhline{1pt}
    \end{tabular}}%
\label{tab:efficiency}%
\end{table*}%

\begin{table*}[t]
\scriptsize
\centering
\caption{The estimated electricity consumption of each method, which is calculated based on the the training and inference runtime. (The computational device is Nvidia A100-40G.)}
\makebox[\linewidth]{
\setlength{\tabcolsep}{0.2em}
\fontsize{9}{9}\selectfont
\begin{tabular}{cccccccc}
    \Xhline{1pt}
    Method & Process & SST-2 & MR & AGNews & R8 & R52 & Avg. Elec Cost \\
    \Xhline{0.75pt}
    \multirow{4}{*}{CARP} & clues and reasoning generating & $11.54kWh$ & $7.29kWh$ & $22.70kWh$ & $1.61kWh$ & $1.46kWh$ \\
                    & KNN with embed. extracting & $0.10kWh$ & $0.06kWh$ & $0.19kWh$ & $0.01kWh$ & $0.01kWh$ \\
                    & result generating & $2.13kWh$ & $26.30kWh$ & $18.53kWh$ & $5.82kWh$ & $6.90kWh$ \\
        \cline{2-7}
                    & total electricity consumption  & $13.77kWh$ & $33.65kWh$ & $41.42kWh$ & $7.44kWh$ & $8.37kWh$ & $20.93kWh$\\
    \Xhline{0.75pt}
    \multirow{3}{*}{LLMEmbed} & embed. extracting & $0.05kWh$ & $1.51kWh$ & $0.21kWh$ & $0.06kWh$ & $0.07kWh$ \\
                                 & downstream model & $0.01kWh$ & $0.004kWh$ & $0.01kWh$ & $0.0009kWh$ & $0.0009kWh$ \\
        \cline{2-7}
                    & total electricity consumption  & $0.06kWh$ & $1.51kWh$ & $0.22kWh$ & $0.06kWh$ & $0.07kWh$ & $0.38kWh$ \\
    \Xhline{1pt}
    \end{tabular}}%
\label{tab:power}%
\end{table*}%

\begin{table*}[t]
\scriptsize
\centering
\caption{The budget comparison between online prompt-based CARP and LLMEmbed.}
\makebox[\linewidth]{
\setlength{\tabcolsep}{0.2em}
\fontsize{9}{9}\selectfont
\begin{tabular}{cccccccc}
    \Xhline{1pt}
    \multicolumn{2}{c}{} & SST-2 & MR & AGNews & R8 & R52 & Avg. budget\\
    \Xhline{0.75pt}
    \multirow{4}{*}{CARP} & tokens of generating clues and reasoning & 22227252 & 41993238 & 32010577 & 2243198 & 4024957 \\
                          & tokens of generating result & 6492145 & 235107723 & 75016765 & 33625747 & 45442485 \\
        \cline{2-7}
                          & total tokens & 28719397 & 277100961 & 107027342 & 35868945 & 49467442 \\
                          & total token consumption & \$57.44 & \$554.20 & \$214.05 & \$71.74 & \$98.93 & \$199.27\\
    \Xhline{0.75pt}
    \multirow{2}{*}{LLMEmbed} & total electricity consumption  & $0.06kWh$ & $1.51kWh$ & $0.22kWh$ & $0.06kWh$ & $0.07kWh$ \\                 
                          & total electricity bill & \$0.0039 & \$0.09815 & \$0.0143 & \$0.0039 & \$0.00455 & \$0.025 \\
    \Xhline{1pt}
    \end{tabular}}%
\label{tab:budget}%
\end{table*}%

\section{Experiment}
\subsection{Setups}
We perform experiments on five widely-used text classification benchmarks, namely
SST-2 \footnote{https://nlp.stanford.edu/sentiment/}, 
MR\footnote{http://www.cs.cornell.edu/people/pabo/movie-review-data/}, 
AGNews\footnote{http://groups.di.unipi.it/$\sim$gulli/AG\_corpus\_of\_news\_articles.html}, 
R8 and R52\footnote{https://www.cs.umb.edu/$\sim$smimarog/textmining/datasets/}. The detailed experimental settings follow CARP \cite{sun2023text}.

\noindent\textbf{SST-2} \cite{socher-etal-2013-recursive} is sampled from snippets of Rotten Tomatoes HTML files. The amounts of train set and test set are 67349 and 872 respectively, and the max length of words is 56.

\noindent\textbf{MR} \cite{maas-EtAl:2011:ACL-HLT2011} contains movie reviews representing positive or negative sentiment. The corpus has 40000 training data and 10000 testing data, of which the max length of words is 2470.

\noindent\textbf{AGNews} \cite{zhang2015character} consists of 4 types of news articles from the AG’s corpus. The dataset contains 120000 training and 7600 testing examples, and the max length of words is 177.

\noindent\textbf{R8} and \textbf{R52} are two subsections of Reuters collection, containing 8 and 52 classifications.
The R8 dataset is composed of 5485 documents for training and 2189 documents for testing, of which the max length of words is 964.
The R52 dataset is composed of 6532 training and 2568 testing documents, of which the max length of words is 1039.

Moreover, we solve these text classification tasks on Nvidia A100-40G platform, and employ LLaMA2-7B \cite{touvron2023llama_b} as the lightweight LLM backbone throughout all of our main experiments. {The batch size we set for each dataset is 1024. The classifier is trained for 100 epochs with initial learning rate $1el^{-4}$.} Our LLMEmbed is compared with recent PLM- and prompt-based baselines \textit{w.r.t.} performance, efficiency and budget.

\subsection{Performance Analysis}
For the conventional PLM method, we employ the widely used BERT and RoBERTa to extract embeddings and adapt them to the downstream text classification tasks. Though the PLM method demonstrates satisfactory performance, there is still scope for enhancement, such as the RoBERTa hardly converges on R52 with only $42.17\%$ accuracy.

For the prompt-based method, we initially leverage the Input-Output (IO) prompting \cite{brown2020language}, which provides input-output pairs as demonstrations to guide the LLM in generating results. Additionally, We implement the SOTA prompt method, CARP \cite{sun2023text}. Under the detailed guidance of CARP, the overall performance of prompt paradigm improve significantly (the average accuracy improve 21.63\%). However, due to the far small parameter scale of LLaMA2(7B) compared to GPT-3(175B), the emergent abilities are somewhat limited. We note that the CARP also hallucinates, and if the prompts exceeds the LLaMA2's input length limitation, CARP will probably generate irrelevant content.

For the LLMEmbed paradigm, we employed fusion strategies as mentioned in Table \ref{tab:fusion}. As shown in Table \ref{tab:acc}, 
it is evident that average pooling of LLM's embeddings, further concatenated with different models' embeddings is most effective. 

\noindent \textbf{Average pooling of LLM's embeddings}: We observe that the overall performance of average pooling surpasses max pooling by 0.1\%$\sim$0.2\% and only-concatenating embeddings by 3\%$\sim$10\%. The average pooling retains more information than max pooling. The only-concatenation degrades the performance due to the too large space of embeddings.

\noindent \textbf{Concatenating with different models' embeddings}: We find that after average pooling of LLM's embeddings, concatenating it with different models' embeddings will further improve the performance, \textit{e.g.} $Avg + Cat$ of three models outperforms two models. This is due to fusing embeddings of generative LLM and discriminative models can complement each other in semantic space. 

\noindent \textbf{Co-occurrence pooling}: This pooling itself may be not the most effective method, but it extract the high-order representation. We concatenate this representation further and get the SOTA performance based on the LLaMA2 backbone. The SOTA LLMEmbed outperforms CARP by 7.34\%, 11.55\%, 10.65\%, 23.12\% and 22.63\% for SST2, MR, AGNews, R8 and R52, respectively. 

Furthermore, we also compare local lightweight LLM-based LLMEmbed with online LLM-based CARP. Despite LLaMA2's significant parameter scale disadvantage compared to GPT-3, the LLMEmbed achieves a comparable performance to the GPT-3 CARP by employing fusion strategies, even outperforms over SST-2, MR, AGNews, R8.

\subsection{Efficiency}
We have measured each process's time cost of employing the prompt paradigm and LLMEmbed paradigm for text classification, and listed the results in Table \ref{tab:efficiency}. Overall, the proposed LLMEmbed paradigm is significantly more efficient than the prompt paradigm.

As shown in Figure \ref{fig:loss}, the downstream model reaches convergence at around 50 epochs. For SST-2, LLMEmbed's time cost is only 0.46\% (00:22:11/79:31:09) of the prompt-based paradigm. Similarly, the ratio of time cost is 3.51\% for MR, 0.48\% for AGNews, 0.66\% for R8 and 0.70\% for R52.

The reasons for the surge in LLMEmbed-based efficiency are: 1) The lightweight LLM only needs to extract the representation of the input texts, without spending time generating answers. 2) LLMEmbed only processes the original input texts without any additional sophisticated prompt words. So the length of input is much smaller than the prompt-based method, resulting in much less computation. 3) LLMEmbed can conduct the text classification in a parallel manner by feeding a batch of texts to the lightweight LLM, whereas prompt-based paradigm can't. We note that if feed a batch of prompts to guide the lightweight LLM to generate results, prompts of this batch will effect each other, leading to a low performance.

As shown in Table \ref{tab:efficiency}, LLMEmbed costs the most time when solving the MR task. Since the length of each text in MR is generally much longer than other datasets, solving MR task requires larger GPU memory, resulting in limited parallel processing. So, the lightweight LLM spends the most time extracting the representation of MR.

\subsection{Budget of Users}

We first compare the electricity consumption between prompt paradigm and LLMEmbed paradigm. For the prompt-based CARP, the power of generating clues and reasoning is about 170$W$, KNN with representation extracting is about 250$W$, and generating result is about 190$W$. So, referring to the time cost listed in Table \ref{tab:efficiency}, the total electricity consumption of SST-2, MR, AGNews, R8 and R52 is  $13.77kWh$, $33.65kWh$, $41.42kWh$, $7.44kWh$ and $8.37kWh$ respectively.

\begin{figure}
\centering
\includegraphics[width=\linewidth]{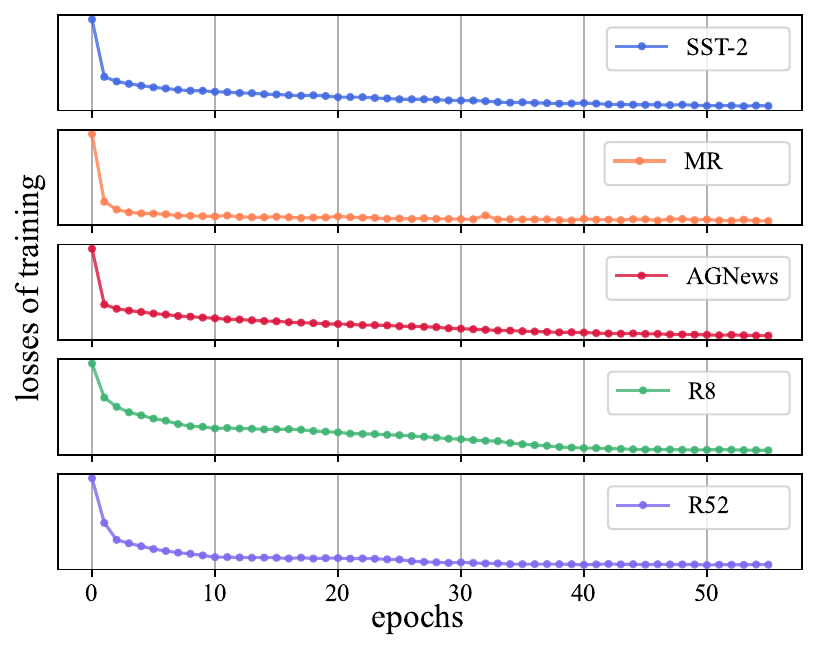}
\caption{The losses of training the downstream model.}
\label{fig:loss}
\end{figure}

For LLMEmbed paradigm, the power of representation extracting is about 240$W$.
When training or testing the downstream model, the power is $35W\sim55W$, which is almost equivalent to the power of a GPU booting up without running any program. We take the average value $45W$ as the power of these processes.
As shown in Figure \ref{fig:loss}, the downstream model converges at around 50 epochs.
Combined with the time cost in Table \ref{tab:efficiency}, the total electricity consumption of SST-2, MR, AGNews, R8 and R52 is $0.06kWh$, $1.51kWh$, $0.22kWh$, $0.06kWh$ and $0.07kWh$ respectively.

The detailed results of electricity consumption on five benchmarks have been listed in Table \ref{tab:power}. For SST-2, LLMEmbed's electricity consumption is merely 0.44\% ($0.06kWh/13.77kWh$) of the prompt-based paradigm. Similarly, the ratio of electricity consumption is 4.49\% for MR, 0.53\% for AGNews, 0.81\% for R8 and 0.84\% for R52. Therefore, it can be seen that LLMEmbed paradigm is greatly more energy-efficient than prompt paradigms.

Following we compare the budget of users between the local LLMEmbed and the online prompt-based LLMs,(\textit{e.g.} GPT\footnote{InstructGPT-3 (text-davinci-003, 175B)}, as CARP does). The pricing of GPT is about \$0.002 per 1$k$ tokens\footnote{https://openai.com/pricing}. We tokenize all the input and output of CARP, and calculate the total sum of these token-based budget. For the local LLMEmbed, its budget is purely decided by the electricity consumption calculated above. Taking the electricity tariffs of Beijing which is \$0.065/$kWh$ as an example, we can easily work out the realistic electricity bill.

All the results have been listed in Table \ref{tab:budget}. Overall, the budget of local LLMEmbed is much smaller than the online prompt-based LLM. The ratio of budget is 0.01\% for SST-2, 0.02\% for MR, 0.01\% for AGNews, 0.01\% for R8 and 0.005\% for R52.

The LLMEmbed paradigm extracts the representations from the original input text, thus obtaining the semantic space through the powerful language comprehension capabilities of LLMs. In this rich semantic space, we can improve the performance of LLMEmbed by employing various semantic representation optimization strategies in text classification. Furthermore, we can adapt LLMEmbed to many other downstream NLP tasks by constructing a proper mapping from the semantic space to the output. Our LLMEmbed is expected to be beneficial for generative LLMs in handling other tasks such as information extraction, script event prediction.

\section{Conclusion}
In this paper, we propose a concise and effective LLM-based paradigm, namely \textbf{LLMEmbed}, to address the text classification task. This novel LLMEmbed paradigm achieves the state-of-the-art performance when solving text classification with lightweight LLMs, even comparable to the performance of LLMs (\textit{e.g.} GPT-3) with sophisticated prompt-based strategies. The LLMEmbed paradigm is flexible and scalable in which the text embeddings obtained from different lightweight LLMs can be fused to improve the overall performance. Moreover, LLMEmbed is far more efficient and budget saving than the prompt-based paradigms. Lastly, we expect that our novel LLMEmbed paradigm can be beneficial for using generative LLMs to handle other down-stream discriminative tasks.

{
\section*{Limitations}
Despite the promising results, LLMEmbed still suffers the following limitations: 1) LLMEmbed needs optimization to further improve the performance in long-text classification tasks if no fine-tuning is performed on the LLM; 2) The explainability of LLMEmbed may be weaker than the prompt-based methods. Our future works will combine the chain-of-thoughts prompts with LLMEmbed to address this issue.
}

\section*{Acknowledgements}
This research was supported by the National Nature Science Foundation of China (No. 62106282) and Beijing Nova Program (No. 20220484139).

\balance
\bibliography{reference}

\end{document}